\documentclass{ecai}
\usepackage{times}
\usepackage{graphicx}
\usepackage{latexsym}
\usepackage{booktabs}
\usepackage{amssymb}
\usepackage{multirow}
\usepackage{todonotes}
\usepackage{hyperref}

\begin{document}

\title{Evaluating Multimodal Representations \\ on Visual Semantic Textual Similarity}

\author{Oier Lopez de Lacalle\institute{University of the Basque Country, Spain, emails: $\{$oier.lopezdelacalle, ander.salaberria, a.soroa, gorka.azcune, e.agirre$\}$@ehu.eus} \and Ander Salaberria$^{1}$\and Aitor Soroa$^{1}$ \and Gorka Azkune$^1$ \and Eneko Agirre$^1$}

\maketitle
\bibliographystyle{ecai}

\begin{abstract}

The combination of visual and textual representations has produced excellent results in tasks such as image captioning and visual question answering, but the inference capabilities of multimodal representations are largely untested.
In the case of textual representations, inference tasks such as Textual Entailment and Semantic Textual Similarity have been often used to benchmark the quality of textual representations. 
The long term goal of our research is to devise multimodal representation techniques that improve current inference capabilities. We thus present a novel task, Visual Semantic Textual Similarity (vSTS), where such inference ability can be tested directly. Given two items comprised each by an image and its accompanying caption, vSTS systems need to assess the degree to which the captions in context are semantically equivalent to each other. Our experiments using simple multimodal representations show that the addition of image representations produces better inference, compared to text-only representations. The improvement is observed both when directly computing the similarity between the representations of the two items, and when learning a siamese network based on vSTS training data. Our work shows, for the first time, the successful contribution of visual information to textual inference, with ample room for benchmarking more complex multimodal representation options.     
\end{abstract}

\section{Introduction}
\label{sec:intro}

Language understanding is a task proving difficult to automatize, because, among other factors, much of the information that is needed for the correct interpretation of an utterance is not explicit in text~\cite{Bruni:2014:MDS:2655713.2655714}. This contrasts with how natural is language understanding for humans, who can cope easily with information absent in text, using common sense and background  knowledge like, for instance, typical spatial relations between objects. From another perspective, it is well-known that the visual modality provides complementary information to that in the text. In fact, recent advances in deep learning research have led the field of computer vision and natural language processing to significant progress in tasks that involve visual and textual understanding. Tasks that include visual and textual content include Image Captioning~\cite{imagenet_cvpr09}, Visual Question Answering~\cite{antol2015vqa}, and Visual Machine Translation~\cite{elliott-etall_VL:2016}, among others.

\begin{figure}[t]
    \centering
    \includegraphics[width=0.48\textwidth]{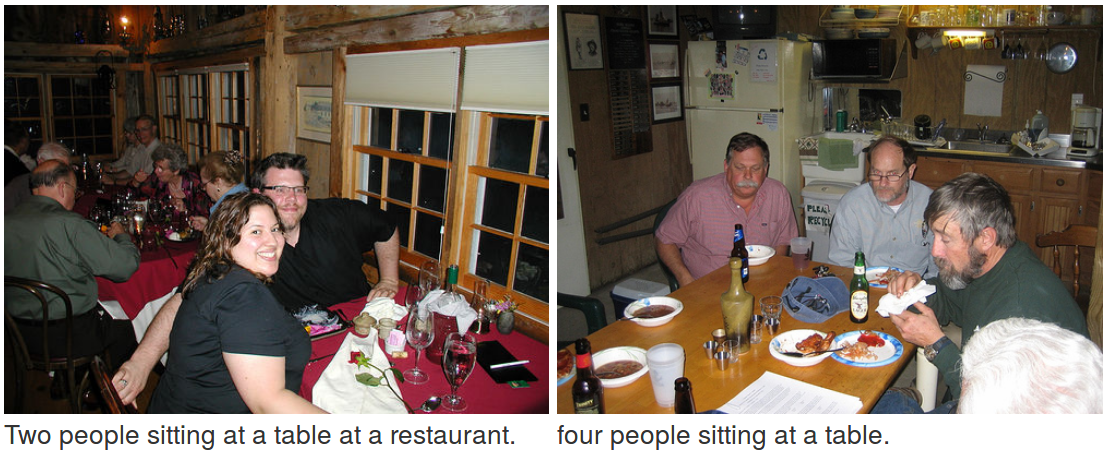}
    \vspace{-0.5cm}\caption{A sample with two items, showing the influence of images when judging the similarity between two captions. While the  similarity for the captions alone was annotated as low (1.8), when having access to the images, the annotators assigned a much higher similarity (4). The similarity score ranges between 0 and 5.  }
    \label{fig:vsts_example}
\end{figure}

On the other hand, progress in language understanding has been driven by datasets which measure the quality of sentence representations, specially those where inference tasks are performed on top of sentence representations, including textual entailment ~\cite{dagan2005pascal,bowman2015large} and semantic textual similarity (STS). In STS~\cite{cer-etal-2017-semeval}, for instance, pairs of sentences have been annotated with similarity scores, with top scores for semantically equivalent sentences and bottom scores for completely unrelated sentences. STS  provides a unified framework for extrinsic evaluation of multiple semantic aspects such as compositionality and phrase similarity. Contrary to related tasks, such as textual entailment and paraphrase detection, STS incorporates the notion of graded semantic similarity between the pair of textual sentences and is symmetric. 

In this paper we extend STS to the visual modality, and present Visual Semantic Textual Similarity (vSTS), a task and dataset which allows to study whether better sentence representations can be built when having access to the corresponding images, in contrast with having access to the text alone. Similar to STS, annotators were asked to score the similarity between two items, but in this case each item comprises an image and a textual caption. Systems need  to predict the human score. Figure~\ref{fig:vsts_example} shows an instance in the dataset, with similarity scores in the captions. The example illustrates the need to re-score the similarity values, as the text-only similarity is not applicable to the multimodal version of the dataset: the annotators return a low similarity when using only text, while, when having access to the corresponding image, they return a high similarity. Although a dataset for multimodal inference exists (visual textual entailment \cite{vu-etal-2018-grounded}) that dataset reused the text-only inference labels.

The vSTS dataset aims to become a standard benchmark to test the contribution of visual information when evaluating the similarity of sentences and the quality of multimodal representations, allowing to test the complementarity of visual and textual information for improved language understanding.
Although multimodal tasks such as image captioning, visual question answering and visual machine translation already show that the combination of both modalities can be effectively used, those tasks do not separately benchmark the inference capabilities of multimodal visual and textual representations.

We evaluate a variety of well-known textual, visual and multimodal representations in supervised and unsupervised scenarios, and systematically explore if visual content is useful for sentence similarity. For text, we studied pre-trained word embeddings such as GloVe~\cite{glove}, pre-trained language models like GPT-2 and BERT ~\cite{devlin2018bert,radford2019language}, sentence representations fine-tuned on an entailment task like USE~\cite{use}, and textual representations pre-trained on a multimodal caption retrieval task like VSE++~\cite{faghri2018vse++}. For image representation we use a model pre-trained on Imagenet (ResNet~\cite{He2015}). 
In order to combine visual and textual representations we used concatenation and learn simple projections. Our experiments show that the  text-only models are outperformed by their multimodal counterparts when adding visual representations, with up to  $24$\% error reduction. 

Our contributions are the following: 
    (1) We present a dataset which allows to evaluate visual/textual representations on an inference task. The dataset is publicly available under a free license\footnote{\url{https://oierldl.github.io/vsts/}}. 
    (2) Our results show, for the first time, that the addition of image representations allows better inference.
    (3) The best text-only representation is the one fine-tuned on a multimodal task, VSE++, which is noteworthy, as it is better than a textual representation fine-tuned in a text-only inference task like USE.
    (4) The improvement when using image representations is observed both when computing the similarity directly from multimodal representations, and also when training siamese networks. At the same time the improvement holds for all textual representations, even those fine-tuned on a similarity task.

\section{Related Work}

The task of Visual Semantic Textual Similarity stems from previous work on textual inference tasks. In textual entailment, given a textual premise and a textual hypothesis, systems need to decide whether the first entails the second, they are in contradiction, or none of the previous ~\cite{dagan2005pascal}. Popular datasets include the Stanford Natural Language Inference dataset ~\cite{bowman2015large}. As an alternative to entailment, STS datasets comprise pairs of sentences which have been annotated with similarity scores. STS systems are usually evaluated on the STS benchmark dataset \cite{cer-etal-2017-semeval}\footnote{See for instance recent models evaluated on STS benchmark \url{http://ixa2.si.ehu.es/stswiki/index.php/STSbenchmark}}. In this paper we present an extension of STS, so we present the task in more detail in the next section. 

Textual entailment has been recently extended with visual information. A dataset for {\bf visual textual entailment} was presented in ~\cite{vu-etal-2018-grounded}. Even if the task is different from the text-only counterpart, they reused the text-only inference ground-truth labels without re-annotating them. In fact, they annotate a small sample to show that the labels change. In addition, their dataset tested pairs of text snippets referring to a single image, and it was only useful for testing grounding techniques, but not to measure the complementarity of visual and textual representations. The reported results did not show that grounding improves results, while our study shows that the inference capabilities of multimodal visual and textual representations improve over text-only representations. 
In related work, \cite{xie2019visual} propose visual entailment, where the premise is an image and the hypothesis is textual. The chosen setting does not allow to test the contribution of multimodal representationn with respect to unimodal ones. 

The complementarity of visual and text representations for improved language understanding was first proven on word representations, where word embeddings were
combined with visual or perceptual input to produce multimodal representations~\cite{frome2013}. The task of Visual Semantic Textual Similarity is also related to other multimodal tasks such as Image Captioning~\cite{bernardi2016automatic,guo2019mscap}, 
Text-Image Retrieval~\cite{barnard2001learning, plummer2015flickr30k} 
and Visual Question Answering~\cite{antol2015vqa}. 

{\bf Image Captioning} is a task that aims to generate a description of a given image. The task is related to ours in that it is required an understanding of the scene depicted in the image, so the system can generate an accurate description of it. Unlike vSTS, image captioning is a generation task in which evaluation is challenging and unclear, as the defined automatic metrics are somewhat problematic \cite{van-miltenburg-etal-2018-measuring}. On the other hand, {\bf Text-Image Retrieval} task requires to find similarities and differences of the items in two modalities, so we can distinguish relevant and irrelevant texts and images regarding the query. Apart from not checking inference explicitly, the other main difference with regards to vSTS is that, in retrieval, items are ranked from most to least similar, whereas the vSTS task consists on scoring an accurate real valued similarity. A comprehensive overview is out of the scope, and thus we focus on the most related vision and language tasks. We refer the reader to~\cite{Mogadala2019TrendsII} for a survey on vision and language research. 

Many of these tasks can be considered as extensions of previously existing NLP taks. For instance, Image Captioning can be seen as an extension of conditional language modeling \cite{de2015survey} or natural language generation~\cite{reiter2000building}, whereas Visual Question Answering is a natural counterpart of the traditional Question Answering in NLP.

Regarding multimodal and unimodal {\bf representation learning}, convolutional neural networks (CNN) have become the standard architecture for generating representations for images~\cite{lecun1995convolutional}. Most of these models learn transferable general image features in tasks such as image classification, and detection, semantic segmentation, and action recognition. Most used transferable global image representations are learned with deep CNN architectures such as AlexNet~\cite{Krizhevsky:2012:ICD:2999134.2999257}, VGG~\cite{Simonyan15}, Inception-v3~\cite{szegedy2016}, and ResNet~\cite{He2015} 
using large datasets such as ImageNet~\cite{imagenet_cvpr09}, MSCOCO~\cite{DBLP:journals/corr/LinMBHPRDZ14} and Visual Genome~\cite{Krishna:2017:VGC:3088990.3089101}. Recently, Graph Convolution Networks (GCN) showed to be promising way to distill multiple input types multimodal representations~\cite{zhang2019graph}.

{\bf Language representation} is mostly done with pretrained word embeddings like Glove~\cite{glove} and sequence learning techniques such as Recurrent Neural Networks (RNN) \cite{Hochreiter:1997:LSM:1246443.1246450}. Recently, self-attention approaches like Transformers~\cite{NIPS2017_7181} provided transferable models (BERT, GPT-2, among others~\cite{devlin2018bert,radford2019language}) that significantly improve many state-of-the-art tasks in NLP. Alternatively, sentence representations have been fine-tuned on an entailment task~\cite{use}. We will present those used in our work in more detail below.

\section{The Visual STS Dataset}

STS assesses the degree to which two sentences are semantically equivalent to each other. The annotators measure the similarity among sentences, with higher scores for more similar sentences. The annotations of similarity were guided by the scale in Table~\ref{tab:sim_def}, ranging from 0 for no meaning overlap to 5 for meaning equivalence. Intermediate values reflect interpretable levels of partial overlap in meaning. 

\begin{table}
    \centering
    \begin{tabular}{l}
    \toprule
        {\bf Similarity definitions:}\\
    \midrule
        5: Completely equivalent: They mean the same thing.\\
        4: Mostly equivalent: Some unimportant details differ.\\
        3: Roughly equivalent: Some important information differs/missing.\\
        2: Not equivalent but share some details.\\
        1: Not equivalent but on the same topic.\\
        0: Completely dissimilar.\\
    \bottomrule
    \end{tabular}
    \caption{Similarity scores with the definition of each ordinal value. Definitions are the same as used in STS datasets~\cite{cer-etal-2017-semeval}}
    \label{tab:sim_def}
\end{table}

In this work, we extend the STS task with images, providing visual information that models use, and assess how much visual content can contribute in a language understanding task. The input of the task now consists of two items, each comprising an image and its corresponding caption. In the same way as in STS, systems need to score the similarity of the sentences with the help of the images.  Figure~\ref{fig:vsts_example} shows an example of an instance in the dataset.

In previous work reported in a non-archival workshop paper \cite{lacallevsts}, we presented a preliminary dataset which used the text-only ground-truth similarity scores.  The 819 pairs were extracted from a subset of the STS benchmark, more specifically, the so called STS-images subset, which contains pairs of captions with access to images from PASCAL VOC-2008~\cite{rashtchian-etal-2010-collecting} and Flickr-8K~\cite{Hodosh:2013:FID:2566972.2566993}. Our manual analysis, including examples like Figure~\ref{fig:vsts_example}, showed that in many cases the text-only ground truth was not valid, so we decided to re-annotated the dataset but showing the images in addition to the captions (the methodology is identical to the AMT annotation method mentioned below). The correlation of the new annotations with regard to the old ones was high (0.9$\rho$) showing that the change in scores was not drastic, but that annotations did differ. The annotators tended to return higher similarity scores, as the mean similarity score across the dataset increased from 1.7 to 2.1. The inter-tagger correlation was comparable to the text-only task, showing that the new annotation task was well-defined. 

From another perspective, the fact that we could only extract 819 pairs from existing STS datasets showed the need to sample new pairs from other image-caption datasets. 
In order to be effective in measuring the quality of multimodal representations, we defined the following desiderata for the new dataset: (1) Following STS datasets, the similarity values need to be balanced, showing a uniform distribution; (2) Paired images have to be different to avoid making the task trivial, as hand analysis of image-caption datasets showed that two captions of the same image tended to be paraphrases of each other; (3) The images should not be present in more than one instance, to avoid biases in the visual side; (4) It has to contain a wide variety of images so we can draw stronger conclusions. The preliminary dataset fulfilled 2 and 3, but the dataset was skewed towards low similarity values and the variety was limited. 

\subsection{Data Collection}
The data collection of sentence-image pairs comprised several steps, including the selection of pairs to be annotated, the annotation methodology, and a final filtering stage.

\paragraph{1. Sampling data for manual annotation.}  We make use of two well-known image-caption datasets. On one hand,  Flickr30K dataset~\cite{Plummer:2017:FEC:3088990.3089104} that has about 30K images with 5 manually generated captions per image. On the other hand, we use the Microsoft COCO dataset~\cite{DBLP:journals/corr/LinMBHPRDZ14}, which contains more than 120K images and 5 captions per image. Using both sources we hope to cover a wide variety of images. 

In order to select pairs of instances, we did two sampling rounds. The goal of the first run is to gather a large number of varied image pairs with their captions which contain interesting pairs. We started by sampling images. 
We then combined two ways of sampling pairs of images. In the first, we generated pairs by sampling the images randomly. This way, we ensure higher variety of paired scenes, but presumably two captions paired at random will tend to have very low similarity. In the second, we paired images taking into account their visual similarity, ensuring the selection of related scenes with a higher similarity rate. We used the cosine distance of the top-layer of a pretrained ResNet-50~\cite{He2015} to compute the similarity of images. We collected an equal number of pairs for the random and visual similarity strategy, gathering, in total, $155,068$ pairs. As each image has 5 captions, we had to select one caption for each image, and we decided to select the two captions with highest word overlap. This way, we get more balanced samples in terms of caption similarity\footnote{We tried random sampling over captions too, but we ended up with a more unbalanced selection.}. 

The initial sampling created thousands of pairs that were skewed towards very low similarity values. Given that manual annotation is a costly process, and with the goal of having a balanced dataset, we used an automatic similarity system to score all the pairs. This text-only similarity system is an ensemble of feature-based machine learning systems that uses a large variety of distance and machine-translation based features. The model was evaluated on a subset of STS benchmark dataset~\cite{cer-etal-2017-semeval} and compared favorably to other baseline models. As this model is very different from current deep learning techniques, it should not bias the dataset sampling in a way which influences current similarity systems.

\begin{figure}
\centering
\includegraphics[scale=0.4]{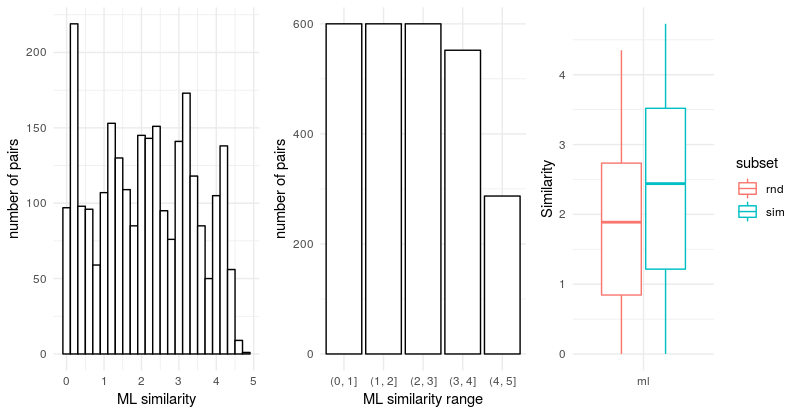}
\caption{Histograms of the similarity distribution in the 2639 sample, according to the automatic text-only system (left and middle plots), and the distribution of the similarity of each sampling strategy ({\bf rnd} stands for random image sampling and {\bf sim} stands for image similarity driven sampling).}
\label{fig:sampling}
\end{figure}

The automatic scores were used to sample the final set of pairs as follows. We defined five similarity ranges ($(0, 1], \ldots ,(4, 5]$) and randomly selected the same amount of pairs from the initial paired sample. We set a sampling of maximum 3000 instances (i.e 600 instances per range). Given the fact that the high similarity range had less than 600 instances, we collected a total of 2639 potential text-image candidate pairs
for manual annotation.
 Figure~\ref{fig:sampling} shows the proposed methodology can sample approximately a uniform distribution with the exception of the higher similarity values (left and middle plots). In addition, we show that the lower predicted similarities are mainly coming from random sampling, whereas, as expected, the higher ones come from similar images.

\paragraph{2. Manual annotations.} In order to annotate the sample of 2639 pairs, we used Amazon Mechanical Turk (AMT). Crowdworkers followed the same instructions of previous STS annotation campaigns~\cite{cer-etal-2017-semeval}, very similar to those in Table \ref{tab:sim_def}. Annotators needed to focus on textual similarity with the aid of aligned images. We got up to 5 scores per item, and we discarded annotators that showed low correlation with the rest of the annotators ($\rho < 0.75$).
In total 56 annotators took part. On average each crowdworker annotated 220 pairs, where the amounts ranged from 19 to 940 annotations. Regardless the annotation amounts, most of the annotators showed high correlations with the rest of the participants. We computed the annotation correlation by aggregating the individual Pearson correlation with averaged similarity of the other annotators. The annotation shows high correlation among the crowdworkers ($\rho = 0.89$ $\pm 0.01$) comparable to that of text-only STS datasets. 

\begin{table}
\centering
\begin{tabular}{lrrrrr}
  \toprule
    & {\bf  \#Pairs} & {\bf Mean} & {\bf Median} & {\bf STD} & {\bf \#Zeros}\\
  \midrule
  Item similarity & 2639 & 1.96 & 1.80 & 1.65 & 549\\
  Item disagreement & 2639 & 0.60 & 0.55 & 0.45 & 724\\
  \bottomrule
\end{tabular}
\caption{Overall item similarity and disagreement of the AMT annotations.}
\label{tab:item_im}
\end{table}

Table~\ref{tab:item_im} shows the average {\bf item similarity} and {\bf item disagreement} in the annotation. We defined item disagreement as the standard deviation of the annotated similarity value. The low average similarity can be explained by the high number of zero-similarity pairs. Item disagreement is moderately low (about 0.6 points out of 5) which is in accordance with the high correlation between the annotators.

\paragraph{3. Selection of difficult examples.} In preliminary experiments, the evaluation of two baseline models, word overlap and the ensemble system mentioned before, showed that the sampling strategy introduced a large number of trivial examples. For example, the word overlap system attained $0.83$ $\rho$. This high correlation could be the result of using word-overlap in the first sampling round. In order to create a more challenging dataset where to measure the effectiveness of multimodal representations, we defined the {\bf easiness} metric to filter out some of the easy examples from the annotated dataset. 

We defined easiness as an amount of discrepancy provided by an example regarding the whole dataset. Taking the inner product of the Pearson correlation formula as basis, we measure the easiness of an annotated example $i$ as follows:
\begin{equation}
    e_{i} = \left(\frac{o_{i} - \overline{o}}{s_{o}} \right) \left(\frac{gs_{i} - \overline{gs}}{s_{gs}} \right)
\end{equation}

where $o_{i}$ is the word-overlap similarity of the $i$-th pair,  $\overline{o}$ is the mean overlap similarity in the dataset, and $s_{o}$ is the standard deviation. Similarly, variable $gs_{i}$ is the gold-standard value of the $i$-th pair, and $\overline{gs}$ and $s_{gs}$ are the mean and standard deviation of gold values in the dataset, respectively. We removed 30\% of the \emph{easiest} examples and create a more challenging dataset of 1858 pairs, reducing $\rho$ to $0.57$ for the word-overlap model, and to $0.66$ $\rho$ (from $0.85$) for the ML based approach. 

\subsection{Dataset Description}

\begin{figure}
    \centering
    \includegraphics[scale=0.4]{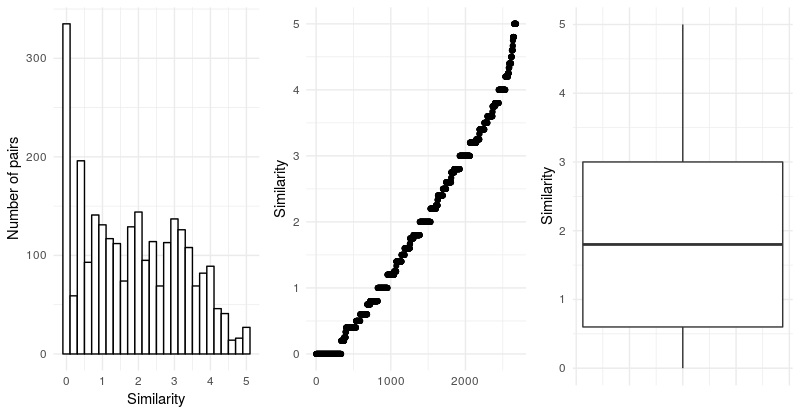}
    \caption{Similarity distribution of the visual STS dataset. Plots show three views of the data. Histogram of the similarity distribution of ground-truth values (left plot), sorted pairs according to their similarity (middle) and boxplot of the similarity values (right). }
    \label{fig:vsts20}
\end{figure}

The full dataset comprises both the sample mentioned above and the 819 pairs from our preliminary work, totalling 2677 pairs. Figure~\ref{fig:vsts20} shows the final item similarity distribution. Although the distribution is skewed towards lower similarity values, we consider that all the similarity ranges are sufficiently well covered. 

Average similarity of the dataset is $1.9$ with a standard deviation of $1.36$ points. The dataset contains 335 zero-valued pairs out of the 2677 instances, which somehow explains the lower average similarity.

\section{Evaluation of Representation Models}
\label{sec:eval}

The goal of the evaluation is to explore whether representation models can have access to images, instead of text alone, have better inference abilities.  We consider the following models.

\textbf{ResNet}~\cite{He2015} is a deep network of 152 layers in which  the residual representation functions are learned instead of learning the signal representation directly. The model is trained over 1.2 million images of ImageNet, the ILSRVC subset of 1000 image categories. We use the
top layer of a pretrained ResNet-152 model to represent the images associated to text. Each image is represented with a vector of 2048 dimensions. 

\textbf{GloVe.} The Global Vector model~\cite{glove} is a log-linear model trained to encode semantic relationships between words as vector offsets in the learned vector space, combining global matrix factorization and local context window methods. Since GloVe is a word-level vector model, we build sentence representations with the mean of the vectors of the words composing the sentence. The pre-trained model from GloVe considered in this paper is the 6B-300d, with a vocabulary of 400k words, 300 dimension vectors and trained on a dataset of 6 billion tokens.

\textbf{BERT.}  The Bidirectional Encoder Representations from Transformer ~\cite{devlin2018bert} implements a novel methodology based on the so-called \textit{masked language model}, which randomly masks some of the tokens from the input, and predicts the original vocabulary id of the masked word based only on its context. The BERT model used in our experiments is the BERT-Large Uncased (24-layer, 1024-hidden, 16-heads, 340M parameters). In order to obtain the sentence-level representation we extract the token embeddings of the last layer and compute the mean vector, yielding a vector of 1024 dimensions.

\textbf{GPT-2.} The Generative Pre-Training-2 model\cite{radford2019language} is a language model based on the transformer architecture, which is trained on the task of predicting the next word, given all the previous words occurring in some text. In the same manner to BERT and GloVe, we extract the token embeddings of the last layer and compute the mean vector to obtain the sentence-level representation of 768 dimensions. The GPT-2 model used in our experiments was trained on a very large corpus of about 40 GB of text data with 1.5 billion parameters.

\textbf{USE.} The Universal Sentence Encoder~\cite{use} is a model for encoding sentences into embedding vectors, specifically designed for transfer learning in NLP. Based on a deep averaging network encoder, the model is trained for varying text lengths, such as sentences, phrases or short textbfs, and in a variety of semantic tasks including STS. The encoder returns the vector of the sentence with 512 dimensions. 

\textbf{VSE++.} The Visual-Semantic Embedding~\cite{faghri2018vse++} is a model trained for image-caption retrieval. The model learns a joint space of aligned images and captions. The model is an improvement of the original introduced by~\cite{DBLP:conf/cvpr/KarpathyL15}, and combines a ResNet-152 over images with a bidirectional Recurrent Neural Network (GRU) over the sentences. Texts and images are projected onto the joint space, obtaining representations of 1024 dimension both for images and texts. 
We used projections of images and texts in our experiments. The VSE++ model used in our experiments was pre-trained on the Microsoft COCO dataset~\cite{DBLP:journals/corr/LinMBHPRDZ14} and the Flickr30K dataset~\cite{Plummer:2017:FEC:3088990.3089104}. 
Table~\ref{tab:models} summarizes the sentence and image representations used in the evaluation.

\begin{table}
    \centering
    \begin{tabular}{lccc}
    \toprule
    {\bf Model} & {\bf Modality} & {\bf dimensions}\\
    \midrule
    {\sc resnet} & Image  & 2048 \\
    {\sc vse++(img)} & Image  & 1024\\
    \midrule
    {\sc glove} & Text  & 300\\ 
    {\sc bert} & Text   & 1024\\
    {\sc gpt-2} & Text  & 768\\
    {\sc use} & Text  &  512\\ 
    {\sc vse++(text)} & Text  & 1024\\
    \midrule
    {\sc concat} & multimodal  & -\\
    {\sc project} & multimodal  & -\\
    \bottomrule
    \end{tabular}
    \caption{Summary of the text and image representation models used.}
    \label{tab:models}
\end{table}

\subsection{Experiments}

\paragraph{Experimental Setting.} We split the vSTS dataset into training, validation and test partitions sampling at random and preserving the overall score distributions. In total, we use 1338 pairs for training, 669 for validation, and the rest of the 670 pairs were used for the final testing. Similar to the STS task, we use the Pearson correlation coefficient ($\rho$) as the evaluation metric of the task. 

\paragraph{STS models.} Our goal is to keep similarity models as simple as possible in order to directly evaluate textual and visual representations and avoid as much as possible the influence of the parameters that intertwine when learning a particular task. We defined two scenarios: the supervised and the unsupervised scenarios.

In the {\bf supervised scenario} we train a Siamese Regression model in a similar way presented in~\cite{tai-etal-2015-improved}. Given a sentence/image pair, we wish to predict a real-valued similarity in some range $[1,K]$, being $K=5$ in our experiments.  We first produce sentence/image representations $h_{L}$ and $h_{R}$ for each sentence in the pair using any of the unimodal models described above, or using a multimodal representations as explained below. Given these representations, we predict the similarity score $o$ using a regression model that takes both the distance and angle between the pair ($h_{L}$, $h_{R}$):

\begin{eqnarray}
    h_{x} &=& h_{L} \odot h_{R}, \\
    h_{+} &=& |h_{L} - h_{R}|,   \\
    h_{s} &=& \sigma (W^{(h)}[h_{x}, h_{+}] + b^{(h)}), \\
    o &=& W^{(o)}h_{s} + b^{(o)}
\end{eqnarray}

Note that the distance and angle concatenation ($[h_{x}, h_{+}]$) yields a $2 * d$-dimensional vector. The resulting vector is used as input for the non-linear hidden layer ($h_{s}$) of the model. Contrary to \cite{tai-etal-2015-improved}, we empirically found that the estimation of a continuous value worked better than learning a softmax distribution over $[1,K]$ integer values. The loss function of our model is the Mean Square Error (MSE), which is the most commonly used regression loss function.

In the {\bf unsupervised scenario} similarity is computed as the cosine of the produced $h_{L}$ and $h_{R}$ sentence/image representations. 

\paragraph{Multimodal representation.} We combined textual and image representations in two simple ways. The first method is concatenation of the text and image representation ({\sc concat}). Before concatenation we applied the L2 normalization to each of the modalities. 
The second method it to learn a common space for the two modalities before concatenation ({\sc project}).

\begin{eqnarray}
    h_{1} &=& \sigma (W^{(1)}m_{1},+ b^{(1)}), \\
    h_{2} &=& \sigma (W^{(2)}m_{2},+ b^{(2)}), \\
    h_{m} &=& [h_{1}, h_{2}]
\end{eqnarray}

The projection of each modality learns a space of $d$-dimensions, so that $h_{1}, h_{2} \in \mathbb{R}^{d}$. Once the multimodal representation is produced ($h_{m}$) for the left and right pairs, vectors are directly plugged into the regression layers.  Projections are learned {\em end-to-end} with the regression layers and the MSE as loss function. 

\paragraph{Hyperparameters and training details.} We use the validation set to learn parameters of the supervised models, and to carry an exploration of the hyperparameters. We train each model a maximum of 300 epochs and apply \emph{early-stopping} strategy with a \emph{patience} of 25 epochs. For early stopping we monitor MSE loss value on validation. For the rest, we run a grid search for selecting the rest of the hyperparameter values. We explore {\em learning rate} values (0.0001, 0.001, 0.01, 0.05),  {\em L2 regularization} weights (0.0, 0.0001, 0.001, 0.01),  and  different hidden layer ($h_{s}$) dimensions (50,100, 200, 300). In addition, we activate and deactivate batch normalization in each layer for each of the hyperparameter selection.

\subsection{Results}
\label{sec:results}

\paragraph{The unsupervised scenario.} Table~\ref{tab:res_uns} reports the results using the item representations directly. We report results over train and dev partitions for completeness, but note that none of them was used to tune the models. As it can be seen, multimodal representations consistently outperform their text-only counterparts. This confirms that, overall, visual information is helpful  in the semantic textual similarity task and that image and sentence representation are complementary. For example, the {\sc bert} model improves more than $13$ points when visual information provided by the {\sc resnet} is concatenated. {\sc glove} shows a similar or even larger improvement, with similar trends for {\sc use} and {\sc vse++(text)}\footnote{{\sc vse++ +resnet} in the table.}.

Although {\sc vse++(img)} shows better performance than {\sc resnet} when applying them alone, further experimentation showed lower complementarity when combining with textual representation (e.g. $0.807\rho$ in test combining textual and visual modalities of {\sc vse++}). This is something expected as {\sc vse++(img)} is pre-trained along with the textual part of the {\sc vse++} model on the same task. We do not show the combinations with {\sc vse++(img)} due to the lack of space. 

Interestingly, results show that images alone are valid to predict caption similarity ($0.627 \rho$ in test). Actually, in this experimental setting {\sc resnet} is on par with {\sc bert}, which is the best purely unsupervised text-only model. Surprisingly, {\sc gpt-2} representations are not useful for text similarity tasks. This might be because language models tend to forget past context as they focus on predicting the next token~\cite{Voita2019TheBE}. Due to the low results of {\sc gpt-2} we decided not to combine it with {\sc resnet}.

\begin{table}
  \centering
  \begin{tabular}{llcccc}
    \toprule
    {\bf Model} & {\bf Modality} & {\bf train $\rho$} & {\bf dev $\rho$} & {\bf test $\rho$} \\
    \midrule
    {\sc glove} & text & 0.576 &  0.580 & 0.587 \\
    {\sc bert} & text  & 0.641 & 0.593  & 0.612 \\
    {\sc gpt-2} & text  & 0.198 & 0.241  & 0.210 \\
    {\sc use} & text & 0.732 & 0.747 & 0.720 \\
    {\sc vse++(text)} & text  &  0.822 &	 {\bf 0.812} & {\bf 0.803}  \\
    \midrule
    {\sc resnet} & image & 0.638 & {0.635} & {0.627}\\
    {\sc vse++(img)}      & image &  0.677 & {\bf 0.666} & {\bf 0.662} \\
    \midrule
    {\sc glove+resnet} & mmodal & 0.736 & 0.732 & 0.730 \\
    {\sc bert+resnet}  & mmodal & 0.768 & 0.747 & 0.745 \\
    {\sc use+resnet}   & mmodal & 0.799 & 0.806 & 0.787 \\
    {\sc vse++ +resnet}& mmodal & 0.846 & {\bf 0.837} & {\bf 0.826} \\
    \bottomrule
  \end{tabular}
 \caption{The unsupervised scenario: train, validation and test results of the unsupervised models.}
 \label{tab:res_uns}
\end{table}

\paragraph{The supervised scenario.} Table~\ref{tab:res_sup}  show a similar pattern to that in the the unsupervised setting. Overall, models that use a conjunction of multimodal features significantly outperform unimodal models, and this confirms, in a more competitive scenario, that adding visual information helps learning easier the STS task. The gain of multimodal models is considerable compared to the text-only models. The most significant gain is obtained when {\sc glove} features are combined with {\sc resnet}. The model improves more than $15.0$ points. In this case, the improvement over {\sc bert} is lower, but still considerable with more than $4.0$ points.

In the same vein as in the unsupervised scenario, features obtained with a {\sc resnet} can be as competitive as some text based models (e.g. {\sc BERT}). {\sc gpt-2}, as in the unsupervised scenario, does not produce useful representations for semantic similarity tasks. Surprisingly, the regression model with {\sc gpt-2} features is not able to learn anything in the training set.
As we did in the previous scenario, we do not keep combining {\sc gpt-2} with visual features.  

Multimodal version of {\sc vse++} and {\sc use}\footnote{{\sc vse++ +resnet} and {\sc use+resnet} models.} are the best model among the supervised approaches. Textual version of {\sc use} and {\sc vse++} alone obtain very competitive results and outperforms some of the multimodal models (the concatenate version of {\sc glove} and {\sc bert} with {\sc resnet}). Results might indicate that text-only with sufficient training data can be on par with multimodal models, but, still, when there is data scarcity, 
multimodal models can perform better as they have more information over the same data point.

Comparison between projected and concatenated models show that projected models attain slightly better results in two cases, but the best overall results are obtained when concatenating {\sc vse++(text)} with {\sc resnet}. Although concatenation proofs to be a hard baseline, we expect that more sophisticated combination methods like grounding \cite{mao16} will obtain larger gains in the future.

\begin{table}
  \centering
  \begin{tabular}{llcccc}
    \toprule
    {\bf Model} & {\bf Modality} &  {\bf train $\rho$} & {\bf dev $\rho$} & {\bf test $\rho$} \\
    \midrule
    {\sc glove} & text & 0.819 & 0.744 & 0.702 \\
    {\sc bert} & text  & 0.888 & 0.775 &  0.781\\
    {\sc gpt-2} & text  & 0.265 & 0.285 & 0.246  \\
    {\sc use} & text & 0.861 & 0.824 & 0.810 \\
    {\sc vse++(text)}  & text & 0.883 & {\bf 0.831} & {\bf 0.825} \\
    \midrule
    {\sc resnet} & image & 0.788 & {\bf 0.721} & {\bf 0.706} \\
    {\sc vse++(img)} & image & 0.775 & 0.703 & 0.701 \\
    \midrule
    {\sc concat: glove+resnet}& mmodal & 0.899 & 0.830 & 0.794 \\
    {\sc concat: bert+resnet} & mmodal  & 0.889 & 0.805 & 0.797 \\
    {\sc concat:  use+resnet} & mmodal & 0.892 & 0.859 & 0.841\\
    {\sc concat: vse++ +resnet} & mmodal & 0.915 & {\bf 0.864} & {\bf 0.852} \\
    \midrule
    {\sc project: glove+resnet} & mmodal & 0.997 & 0.821 & 0.826\\
    {\sc project: bert+resnet} & mmodal  & 0.996 & 0.825 & 0.827 \\
    {\sc project: use+resnet} & mmodal & 0.998 & 0.850 & 0.837\\
    {\sc project: vse++ +resnet} & mmodal & 0.998 & {\bf 0.853} & {\bf 0.847} \\
    \bottomrule
  \end{tabular}
   \caption{Supervised scenario: Train, validation and test results of the unsupervised models}
 \label{tab:res_sup}
\end{table}

\section{Discussion}
\subsection{Contribution of the Visual Content}

Table~\ref{tab:contribution} summarizes the contribution of the images on text representations in test partition. The contribution is consistent through all text-based representations. We measure the absolute difference ({\bf Diff}) and the error reduction ({\bf E.R}) of each textual representation with the multimodal counterpart. For the comparison we chose the best text model for each representation.  As expected we obtain the largest improvement  ($22-26\%$ E.R) when text-based unsupervised models are combined with image representations. Note that unsupervised models are not learning anything about the specific task, so the more information in the representation, the better. In the case of {\sc use} and {\sc vse++} the improvement is significant but not as large as the purely unsupervised models.
The best text-only
representation is the one fine-tuned on a multimodal task, VSE++,
which is noteworthy, as it is better than a textual representation fine-tuned in a text-only inference task like USE.

Improvement is consistent for the supervised models. Contrary to the unsupervised setting, these models are designed to learn about the task, so there is usually less room for the improvement. Still, {\sc glove+resnet} shows an error reduction of $12.9$ in the test set.  Finally, {\sc use} and {\sc vse++}  show smaller improvements when we add visual information into the model. 
\begin{table}
\centering
  \begin{tabular}{llllcc}
    \toprule
    {\bf Scenario} & {\bf Repr} & {\bf text} & {\bf mmodal} & {\bf Diff} & {\bf E.R} \\
    \midrule
    Unsup & {\sc glove}  & 0.587 & 0.730 & 0.143 & 24.4 \\
    Unsup & {\sc bert}   & 0.612 & 0.745 & 0.133 & 21.7 \\
    Unsup & {\sc use}    & 0.720 & 0.787 & 0.067 & 9.3  \\
    Unsup & {\sc vse++}    & 0.803 & 0.826 & 0.023 & 2.9 \\
    \midrule
    Sup & {\sc glove}    & 0.702 & 0.793 & 0.091 & 12.9 \\
    Sup & {\sc bert}     & 0.781 & 0.827 & 0.046 & 5.8  \\
    Sup & {\sc use}      & 0.810 & 0.841 & 0.031 & 3.8  \\
    Sup & {\sc vse++}      & 0.825 & 0.852 & 0.027 & 3.3 \\
    \bottomrule
  \end{tabular}
  \caption{Contribution of images over text representations on test.}
  \label{tab:contribution}
\end{table}

\begin{figure}[t]
    \centering
    \includegraphics[scale=0.31]{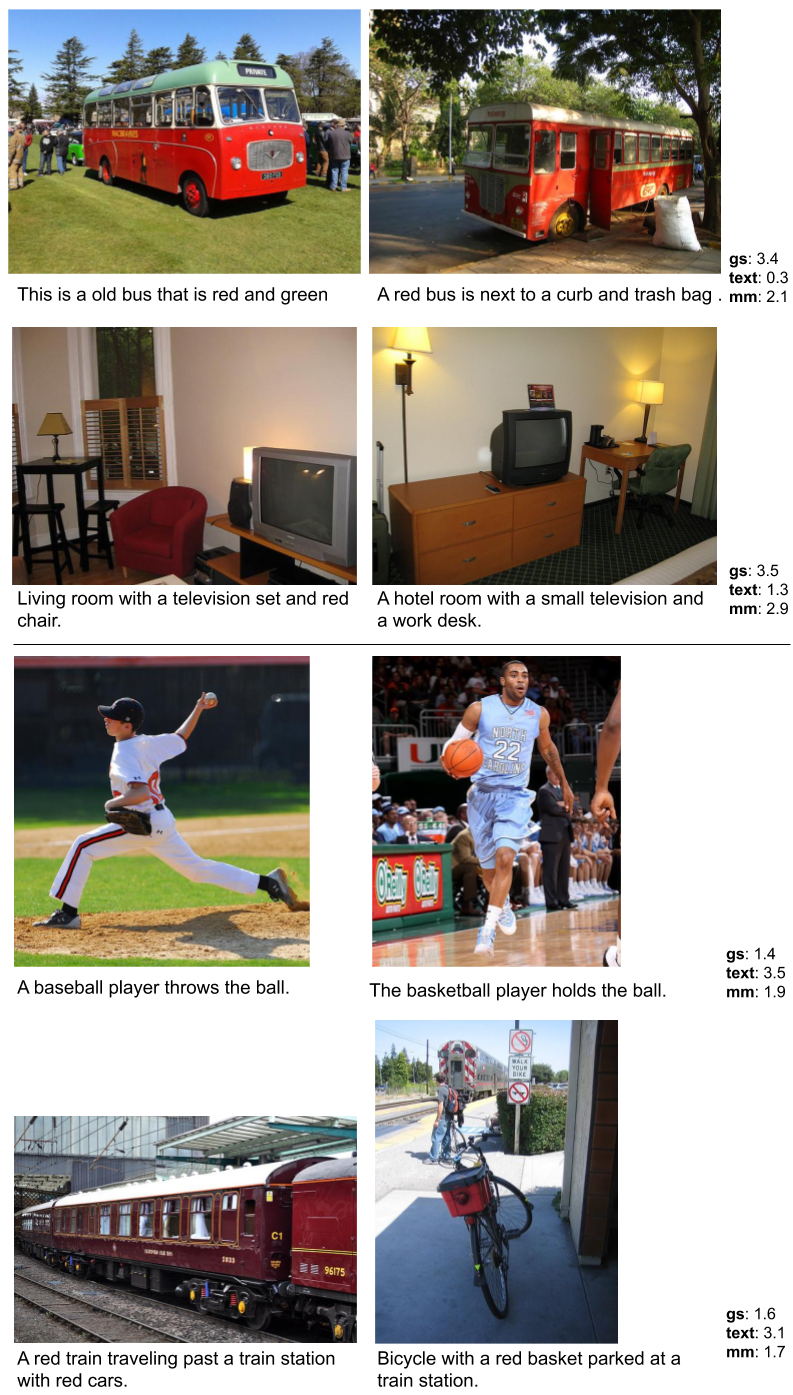}
    \caption{Examples of the contribution of the visual information in the task. {\bf gs} for gold standard similarity value, {\bf text} and {\bf mm} for text-only and multimodal models, respectively. On top examples where related descriptions are lexicalized differently and images help. On the bottom cases where similar words are used to describe different situations.}
    \label{fig:image_contribution}
\end{figure}

Figure \ref{fig:image_contribution} displays some examples where visual information positively contributes predicting accurately similarity values. Examples show the case where related descriptions are lexicalized in a different way so a text-only model ({\sc glove}) predicts low similarity between captions (top two examples). Instead, the multimodal representation {\sc glove+resnet} does have access to the image and can predict more accurately the similarity value of the two captions. The examples in the bottom show the opposite case, where similar set of words are used to describe very different situations. The text based model overestimates the similarity of captions, while the multimodal model corrects the output by looking at the differences of the images. 

\begin{figure}[!t]
    \centering
    \includegraphics[scale=0.31]{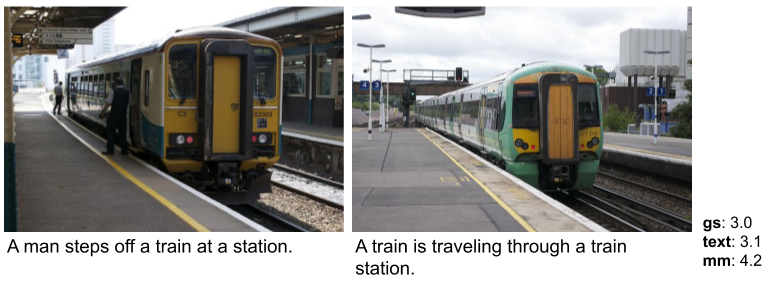}
    \caption{Example of misleading images. The high similarity of images makes the prediction of the multimodal model inaccurate, while the text only model focuses on the most discriminating piece of information. Note that {\bf gs} refers to the gold standard similarity value, and {\bf text} and {\bf mm} refer to text-only and multimodal models, respectively.}
    \label{fig:misleading}
\end{figure}

On the contrary, Figure~\ref{fig:misleading} shows that images can also be misleading, and that the task is not as trivial as combining global representations of the image. In this case, related but different captions are supported by very similar images, and as a consequence, the multimodal model overestimates their similarity, while the text-only model focuses on the most discriminating piece of information in the text.

\subsection{The effect of hyperparameters}
\begin{figure}[h]
\includegraphics[scale=0.7]{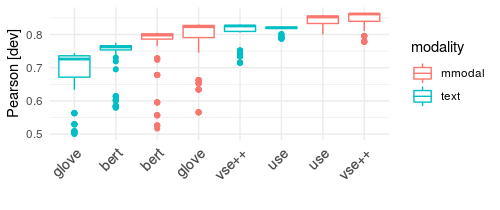}
\caption{Variability of the supervised models regarding hyperparameter selection on development. The multimodal models use concatenation. Best viewed in colour. }
\label{fig:var}
\end{figure}

Neural models are sensitive to hyperparameters, and we might think that results on the supervised scenario are due to hyperparameter optimization. Figure~\ref{fig:var} displays the variability of $\rho$ in development across all hyperparameters.  Due to space constraints we show text-only and multimodal concatenated models. Models are ordered by mean performance. As we can see, combined models show better mean performance, and all models except Glove exhibit tight variability.  


\section{Conclusions and Future Work}
\label{sec:conclusions}
The long term goal of our research is to devise
multimodal representation techniques that improve current inference
capabilities. We have presented a novel task, Visual Semantic Textual
Similarity (vSTS), where the inference capabilities of visual, textual, and multimodal representations can be tested directly. The dataset has been manually annotated by crowdsourcers with high inter-annotator correlation ($\rho = 0.89$). We tested several well-known textual and visual representations, which we combined using concatenation and projection.  Our results show, for the first time, that the addition of
image representations allows better inference. The best text-only
representation is the one fine-tuned on a multimodal task, VSE++,
which is noteworthy, as it is better than a textual representation fine-
tuned in a text-only inference task like USE. The improvement
when using image representations is observed both when computing the similarity directly from multimodal representations, and also
when training siamese networks.

In the future, we would like to ground the text representations to image regions ~\cite{mao16}, which could avoid misleading predictions due to the global representation of the image. Finally, we would like to extend the dataset with more examples, as we acknowledge that training set is limited to train larger models. 

\ack

This research was partially funded by the Basque Government excellence research group (IT1343-19), the NVIDIA GPU grant program, the Spanish MINECO (DeepReading RTI2018-096846-B-C21 (MCIU/AEI/FEDER, UE)) and project BigKnowledge (Ayudas Fundación BBVA a equipos de investigación científica 2018). Ander enjoys a PhD grant from the Basque Government.

\bibliography{ecai}
\end{document}